\documentclass[10pt, a4paper]{article}
\usepackage{lrec}

\title{WAC: A Corpus of Wikipedia Conversations for Online Abuse Detection}

\name{Noé Cécillon, Vincent Labatut, Richard Dufour and Georges Linarès}

\address{Laboratoire Informatique d'Avignon -- LIA EA 4128, Avignon Université, France \\
         \{firstname.lastname\}@univ-avignon.fr\\}

\abstract{
With the spread of online social networks, it is more and more difficult to monitor all the user-generated content. Automating the moderation process of the inappropriate exchange content on Internet has thus become a priority task. Methods have been proposed for this purpose, but it can be challenging to find a suitable dataset to train and develop them. This issue is especially true for approaches based on information derived from the structure and the dynamic of the conversation.
In this work, we propose an original framework, based on the Wikipedia Comment corpus, with comment-level abuse annotations of different types. The major contribution concerns the reconstruction of conversations, by comparison to existing corpora, which focus only on isolated messages ({\it i.e.} taken out of their conversational context). This large corpus of more than 380k annotated messages opens perspectives for online abuse detection and especially for context-based approaches. 
We also propose, in addition to this corpus, a complete benchmarking platform to stimulate and fairly compare scientific works around the problem of content abuse detection, trying to avoid the recurring problem of result replication. Finally, we apply two classification methods to our dataset to demonstrate its potential. \\ 
\newline \Keywords{Wikipedia conversations, Abuse detection, Evaluation framework, Automatic moderation} }

\begin{document}

\maketitleabstract

\section{Introduction}
The ever growing quantity of content posted online requires more and more moderators to monitor this content. A fast and accurate moderation is highly beneficial to online platforms, but it is increasingly expensive and difficult to maintain. Therefore, the automated detection of abusive online content is an important research topic.
Corpora allowing to develop such methods often focus on \textit{single comments} without any conversational context~\cite{Pavlopoulos2017,Razavi2010}.
Yet, recent works~\cite{Papegnies2019,Yin2009} suggest that considering the \textit{entire conversation} thread might improve the automatic detection of abusive content. However, the development of such methods is currently limited by the lack of large scale corpora of conversations. Corpora containing full conversations exists but they have limited number of messages or are not publicly available~\cite{Napoles2017,Cecillon2019}.
\newcite{Karan2019} offers a solution with \textit{PreTox}, a large corpus of discussion threads from Wikipedia talk pages. However, the quality of their semi-automatically generated annotations might be problematic, as the authors report a Precision of only $51$\%.

In this paper, we reconstruct a large scale corpus of messages  from English Wikipedia talk pages, structured as full conversations and annotated with high quality annotations. This results in a corpus containing roughly 193k conversations and 383k messages annotated as being abusive or not. To encourage further development of context- and thread-based methods in the area of abusive content detection, we publish and make freely available this corpus and the code source used for its extraction. 
The other objective of this work is to improve replicability and ease the comparison of classification methods. In this context, we introduce an open-source benchmarking platform that we developed.

The contribution of this work is threefold. First, we match two existing corpora of Wikipedia messages and develop a pipeline to create a large publicly available corpus of conversations. Messages are provided together with detailed information such as the message type, author, talk page and high quality annotations. Second, we present a common comparison platform grouping approaches and methods to stimulate communities around automatic detection of abusive content. Third, we illustrate the interest of our corpus and platform by assessing existing abuse detection methods.

The rest of this article is organized as follows. First, in Section~\ref{sec:RelatedWork}, we describe the existing corpora related to our proposed one, and how they are used in the literature. Then, we describe our corpus in Section~\ref{sec:ProposedCorpus}, as well as the reconstruction pipeline we propose for its constitution. We present a benchmarking platform and some results we obtain on our corpus in Section~\ref{sec:ProposedBenchmark}. Finally, in Section~\ref{sec:Conclusion}, we summarize our results and present some perspectives.

\section{Related Work}
\label{sec:RelatedWork}
In this section, we introduce the corpora of Wikipedia messages related to abuse detection. We review how they are used in the literature, and stress the limitations of these corpora as well as the works leveraging them.

\subsection{Wikipedia Talk Pages} 
\label{subsec:WPtp}
\begin{figure}[!ht]
    \begin{center}
        \includegraphics[width=\columnwidth]{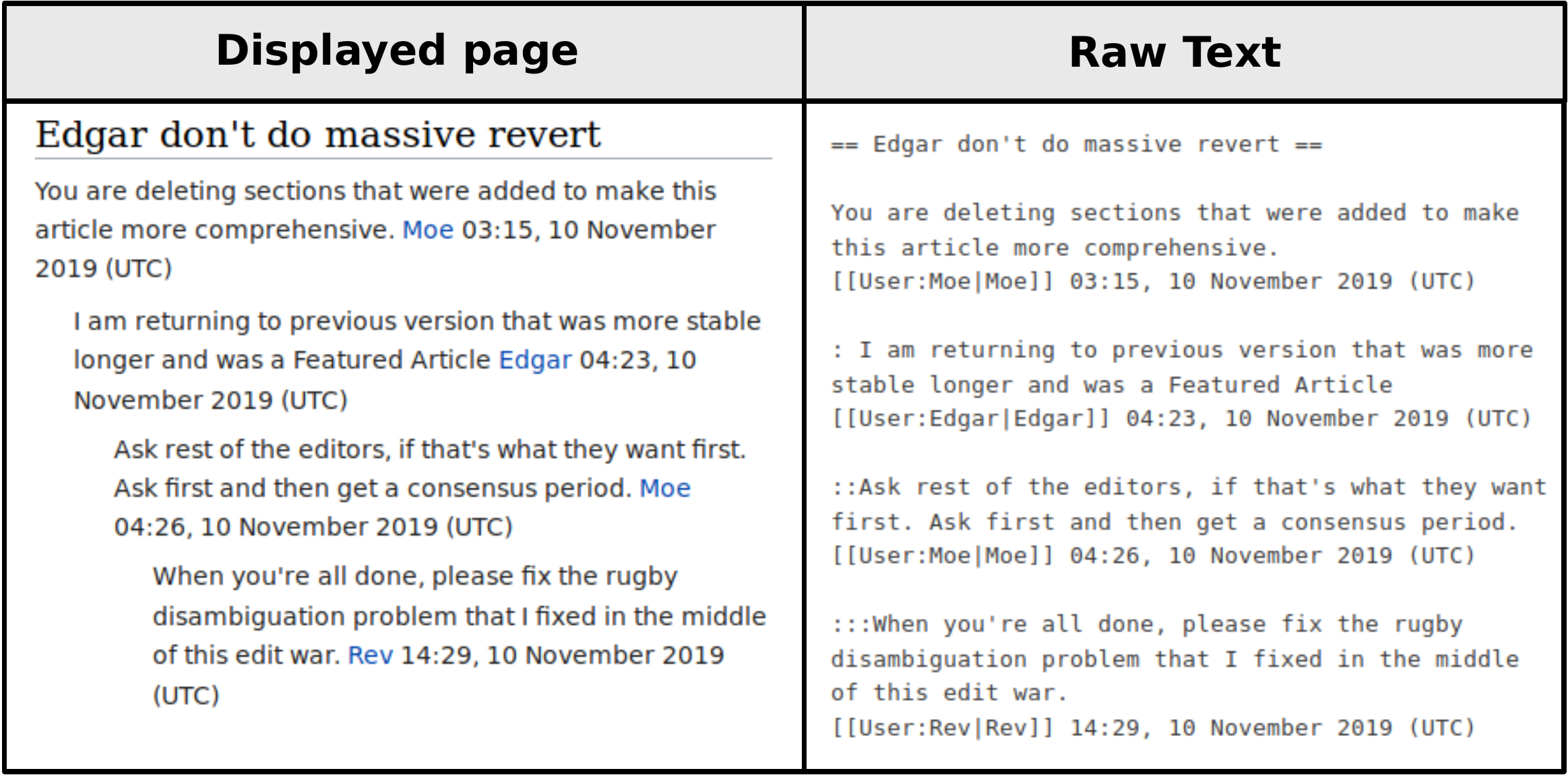} 
        \caption{Part of the \textit{Japan} Wikipedia article  talk page: rendered page (left) and corresponding Wikicode (right).}
        \label{fig:ConventionFormat}
    \end{center}
    \vspace{-2mm}
\end{figure}

A \textit{talk page} is a discussion page where users can argue and discuss topics relative to a specific Wikipedia page. Every Wikipedia user and article has a related talk page, identified by a unique \texttt{page\_id}. But Wikipedia does not propose a standard post system such as those commonly used in online forums. Instead, the talk page is similar to a regular Wikipedia article page, or a wiki page in general: in theory, users have the ability to edit it by adding, modifying or removing text anywhere. However, in practice, a set of writing and formatting conventions\footnote{\label{note:convention}\url{https://en.wikipedia.org/wiki/Help:Talk_pages\#Replying_to_an_existing_thread}} allow giving structure to the various conversations taking place on the talk page. For instance, when a user adds his own post, he indents it so as to indicate its hierarchical level in the conversation tree. Figure~\ref{fig:ConventionFormat} shows an example of Wikipedia conversation under the form of the rendered talk page and the corresponding Wikicode (Wikipedia markup language). Note that a talk page generally contains several conversations at once.

Like for article pages, Wikipedia stores the changes corresponding to each edit, as a \textit{revision} entry containing the text of the page after the edit. Each revision is identified by a unique number called \texttt{rev\_id}.

\subsection{Wikipedia Comment Corpus} 
\label{subsec:WCC}
As part of the Wikipedia Detox research project, \newcite{Wulczyn2017} proposed the \textit{Wikipedia Comment Corpus} (WCC), a corpus of discussion comments from English Wikipedia talk pages. These comments are extracted using the revision history of each considered talk page. The authors consider the textual differences between two consecutive revisions of the talk page, and distinguish two cases depending on the importance of these changes. If the modification is significant, they assume a new comment was posted, which is identified by its own \texttt{rev\_id}. Otherwise, they suppose an existing comment was modified, and apply these changes without updating its \texttt{rev\_id}. Therefore, a given \texttt{rev\_id} is associated to \textit{at most} one comment, and one comment to \textit{exactly} one \texttt{rev\_id}. However, it is important to note that one large edition can correspond, in practice, to a user writing \textit{several} new posts in distinct conversations of the same page. In this case, all these posts are mistakenly gathered in a single WCC comment.

\newcite{Wulczyn2017} used a public dump of the English Wikipedia full history made available in January 2016 to create their corpus, which contains more than 63M comments posted between 2004 and 2015. From this massive corpus, they sampled 3 smaller datasets that they annotated for different types of abuse: 
\begin{itemize}
    \item \textit{personal attack}: abusive content directed at somebody's person rather than providing evidence;
    \item \textit{aggression}: malicious remark to a person or group on characteristics such as religion, nationality or gender;
    \item \textit{toxicity}: comment that can make other people want to leave the conversation.
\end{itemize}

It is important to note that each comment in the three datasets is \textit{explicitly} annotated as abusive or not. By comparison, in the abuse detection literature, datasets are often annotated by considering comments flagged by moderators as abusive, whereas the rest of the comments are deemed non-abusive \textit{by default}, without further check. It is then possible for the non-abusive label to be assigned to {\it abusive} comments just because they were missed by the human moderators \textit{e.g.}~\cite{Papegnies2017,Delort2011,Karan2019}. Having explicitly annotated non-abusive comments makes WCC a more reliable corpus, on this aspect.

Information on the datasets is summarized in Table~\ref{tab:InfoCorpus}. The \textit{Personal attack} and \textit{Aggression} datasets contain exactly the same 115k comments while the \textit{Toxicity} dataset contains more comments (159k). Among them, 77k appear in all three datasets. The prevalence of abusive comments in the \textit{Personal attack} dataset is 13.4\%, 14.7\% in the \textit{Aggression} dataset, and 11.5\% in the \textit{Toxicity} dataset. This prevalence does not reflect the data though, as \newcite{Wulczyn2017} oversampled comments from blocked users to enhance the variety of abusive comments. The original abuse rate in Wikipedia comments is around 1\%.

\begin{table}[!ht]
\centering
  \begin{tabularx}{\columnwidth}{|X|r|r|l|}
    \hline
    \textbf{Dataset} & \textbf{Comments} & \textbf{Percentage} & \textbf{Type of} \\
     &  & \textbf{abusive} & \textbf{annotation} \\
    \hline
    \textbf{Personal} & 115,864 & 13.4 \% & binary \\
    \textbf{attack} & & & \\
    \hline
    \textbf{Aggression} & 115,864 & 14.7 \% & binary and \\
    & & & numerical \\
    \hline
    \textbf{Toxicity} & 159,686 & 11.5 \% & binary and \\
    & & & numerical \\
    \hline
  \end{tabularx}
  \caption{Main properties of the three datasets constituting the Wikipedia Comment Corpus (WCC).}
  \label{tab:InfoCorpus}
\end{table}

As the \textit{Wikipedia Comment Corpus} (WCC) is one of the largest available human annotated comments corpus, it is used in many works. \newcite{Wulczyn2017} themselves tackle the problem of detecting personal attacks in Wikipedia comments. They experiment with logistic regression and multi-layer Perceptron classifiers using word or character $n$-gram features and report a $96.59$ AUC score on the \textit{Personal attack} dataset. \newcite{Pavlopoulos2017} apply deep learning methods to the moderation task. They experiment with various methods such as Convolutional Neural Network (CNN) operating on word embeddings, Recurrent Neural Network (RNN) and several variants of RNN using an attention mechanism. Their results on the \textit{Personal attack} and \textit{Toxicity} datasets outperform previously reported results with an AUC score up to $98.42$ on the \textit{Toxicity} dataset. \newcite{Grondahl2018} propose a comparative analysis of state-of-the-art hate speech detection models and apply them to the \textit{Personal attack} dataset. They experiment with Logistic Regression (LR) and Multi-Layer Perceptrons (MLP) operating on character $n$-grams, Convolutional Neural Networks (CNN) and Long Short-Term Memory (LSTM) approach. The report results ranging from $85$ to $87$\% in terms of macro-averaged $F1$-score. \newcite{Mishra2018} propose various methods to detect abusive content in the \textit{Personal attack} and achieve their best results with a method using context-aware representations for characters. This method is referred to as the {\it Context hidden-state + char $n$-grams} method, and obtains a $89.35$ and $87.44$ macro-averaged $F1$-score on the \textit{Toxicity} and \textit{Personal attack} datasets, respectively.

Furthermore, \newcite{Dixon2018} show that classifiers can have unfair biases toward certain people and propose methods to mitigate unintended bias in text classification models. They illustrate this statement by applying their methods to the \textit{Toxicity} and \textit{Personal attack} datasets. 

Table~\ref{tab:RecapPerformances} summarizes the performances reported in the previously cited articles.  The second column refers to the datasets used to train and test the classifier. Additionally, we can see that the \textit{Aggression} dataset is not used by any of the listed methods. The third column indicates, when available, the percentage of comments in each of the train/development/test split. 

Although the classification performances reported are quite good, most of these models can be fooled by basic obfuscation or adversarial methods. \newcite{Hosseini2017} demonstrate the efficiency of such an attack against the \textit{Google Perspective} API. \newcite{Grondahl2018} present some basic, but efficient, evasion methods. Most of them induce a significant decrease in the classification performances. The  word-based are the most vulnerable ones, which can be completely fooled by introducing or removing manual typos, punctuation and spaces in comments. Because of this possible vulnerability, it can be interesting to rely on more information than only the textual content of each comment.

\begin{table*}[!ht]
\centering
  \begin{tabular} { |m{7.9cm}|m{3.4cm}|l|l|l| }
    \hline
    \textbf{Method} & \textbf{Dataset} & \textbf{Split} & \textbf{Metric} & \textbf{Score} \\
    \hline
    Logistic Regression \cite{Wulczyn2017} & Personal attack & 60/20/20 $^*$ & AUC & 96.24 \\
    \hline
    Multi-layer perceptrons \cite{Wulczyn2017} & Personal attack & 60/20/20 $^*$ & AUC & 96.59 \\
    \hline
    RNN \cite{Pavlopoulos2017} & Toxicity & N/A & AUC & 98.42 \\
    \hline
    RNN + attention  mechanism \cite{Pavlopoulos2017} & Personal attack & N/A & AUC & 97.46 \\
    \hline
    RNN + attention  mechanism \cite{Pavlopoulos2017} & Toxicity/Personal attack & N/A & AUC & 98.22 \\
    \hline
    LSTM \cite{Grondahl2018} & Personal attack & N/A & F1-score & 85. \\
    \hline
    CNN + GRU \cite{Grondahl2018} & Personal attack & N/A & F1-score & 87. \\
    \hline
    Context hidden-state+char $n$-grams \cite{Mishra2018} & Personal attack & 60/40 $^+$ & F1-score & 87.44 \\
    \hline
    Context hidden-state+char $n$-grams \cite{Mishra2018} & Toxicity & 60/40 $^+$ & F1-score & 89.35 \\
    \hline
  \end{tabular}
  \caption{Works leveraging the WCC, with the obtained performances. The \textit{Split} column corresponds to the percentage of comments in the train/test or train/development/test sets. Symbols ($^*$,$^+$) denotes a similar split used by several methods.}.
  \label{tab:RecapPerformances}
  \vspace{-4mm}
\end{table*}

\subsection{WikiConv}
\label{subsec:WikiConv}
\textit{WikiConv}~\cite{Hua2018} is a large public corpus based on Wikipedia talk pages extracted from a July 2018 Wikipedia dump. This corpus contains \textit{full conversations}, and not only \textit{isolated comments} like WCC. In this corpus, we call ''messages'' the textual elements constituting a conversation.
The structure of a conversation is retrieved by considering the revision history of the talk page containing it. This history is viewed as a sequence of \textit{conversational actions}. A conversational action is an object representing one operation performed by a user on a talk page. It is composed of many attributes about the action, the talk page, the conversation and its structure. Additionally, actions are categorized into 5 types: conversation thread \textit{creation}, new message \textit{addition}, existing message \textit{modification}, message \textit{deletion}, and deleted message \textit{restoration}. All the attributes are listed and described on the WikiConv authors' GitHub repository\footnote{\href{https://github.com/conversationai/wikidetox/tree/master/wikiconv}{https://github.com/conversationai/wikidetox/tree/master/wikiconv}}. As mentioned before, when performing these actions, the Wikipedia users respect a set of formatting conventions. Hua \textit{et al}. define a heuristic leveraging this knowledge to identify actions, retrieve their description, and determine their type. This pipeline only relies on visual markup clues, so it is language-independent and can be applied to any version of Wikipedia archives, as long as the formatting conventions stay the same. The largest component of the corpus is from the English Wikipedia but the pipeline is also applied to Chinese, Russian, Greek and German Wikipedia.

\begin{figure}[!ht]
    \begin{center}
        \includegraphics[width=0.93\columnwidth]{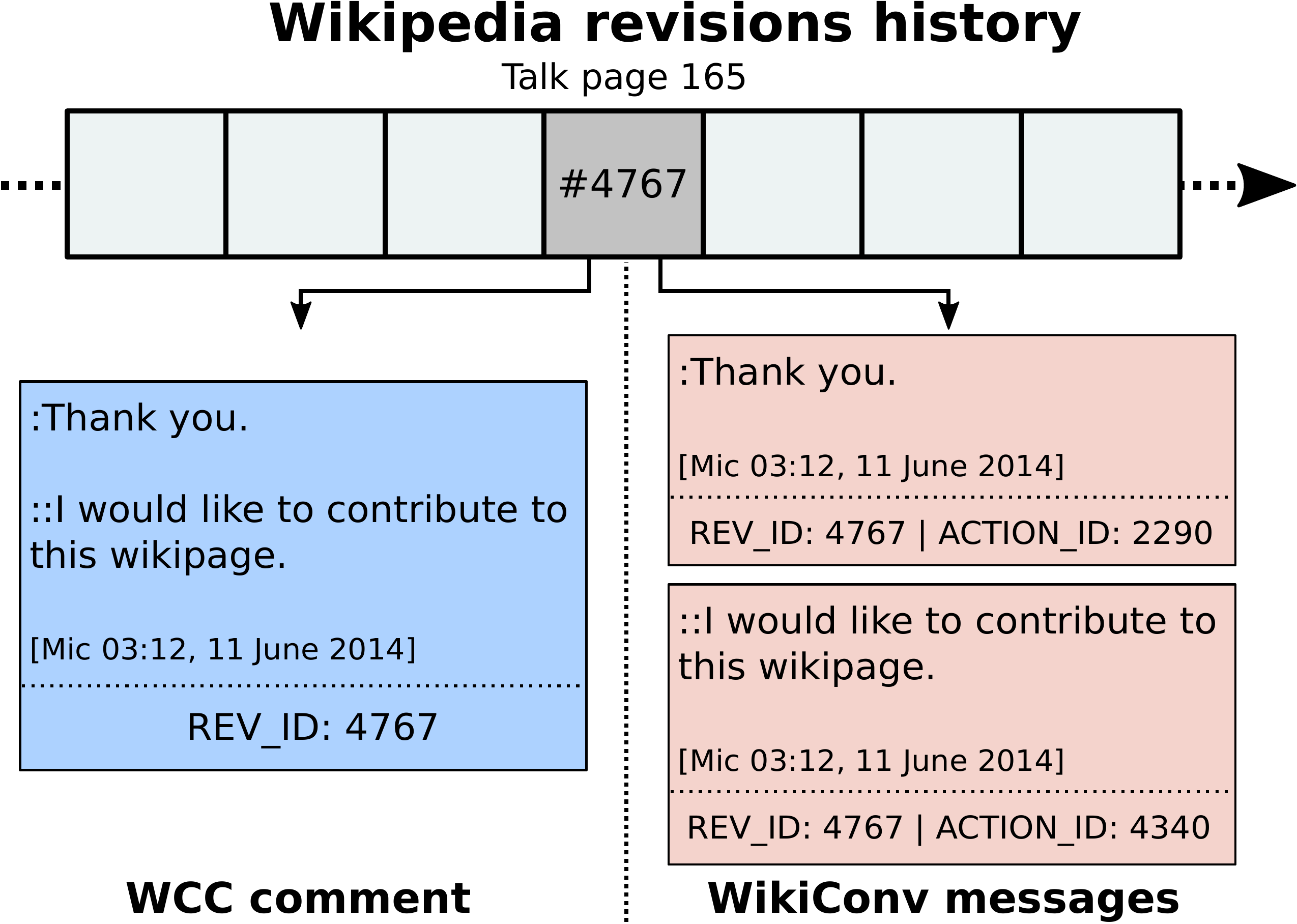} 
        \caption{Illustration of the issue regarding the \texttt{rev\_id} between the WCC and WikiConv.  Figure available at \href{https://doi.org/10.6084/m9.figshare.11302385}{10.6084/m9.figshare.11302385} under CC-BY license.}
        \label{fig:RevidExample}
    \end{center}
\end{figure}

A WikiConv message is the textual content associated to an action, i.e. the text that was added, removed, or edited. A single revision of a talk page can be constituted of several actions, and therefore result in several WikiConv messages. Because of that, the \texttt{rev\_id}, which is used as a unique comment identifier in WCC, can be shared by multiple actions in WikiConv. Instead, a WikiConv message is uniquely identified by an \texttt{action\_id}. Another major difference with WCC is that WikiConv contains the full history of the conversation, with each successive version of a message in case it is edited, and not only its final form. Moreover, when several new posts are added in one revision, those are not merged in a single comment as in WCC, but represented by separate WikiConv messages. It is therefore quite common that multiple WikiConv messages correspond to the same WCC comment. Figure~\ref{fig:RevidExample} illustrates a typical example where the added text is split over two different levels of indentation. In WCC, all the text is concatenated into a single comment while in WikiConv, $2$ messages (and therefore actions) are created, each corresponding to a different indentation level. The $2$ actions are distinct (different \texttt{action\_id}) but they have the same \texttt{rev\_id}. 
Finally, the most important difference with WCC is that the messages are not annotated for abuses.

Moreover,~\newcite{Hua2018} use the Google Perspective API\footnote{\href{https://www.perspectiveapi.com}{https://www.perspectiveapi.com}} to score the toxicity of the messages in WikiConv. Based on machine learning models, this API scores messages for several types of abuse, \textit{e.g.}, toxicity, profanity, threat, insult. In WikiConv, all messages are scored on their toxicity and severe toxicity. These $2$ are provided as attributes of each message in the corpus.

This corpus is quite recent, so only few researches currently  use it. We report one published paper by \newcite{Chang2019}, that explores the potential of a small subset of conversations sourced from WikiConv to predict future derailment in online conversations. In the next section, we introduce a corpus extending the work of \newcite{Hua2018} on WikiConv.

\subsection{PreTox}
\label{subsec:PreTox}
In a recent work, \newcite{Karan2019} proposed \textit{PreTox}, a corpus based on WikiConv (see previous section). \textit{PreTox} is composed of complete discussion threads with semi-automatically generated toxicity annotations. Karan \textit{et al.} rely on a heuristic to flag toxic messages. This heuristic combines two types of information: 1) whether or not the message was deleted by someone else; and 2) the scores generated using the Google Perspective API and provided along with the WikiConv corpus. A binary toxicity annotation is created for each message using this heuristic. Flagged messages are deemed toxic and, unlike WCC, all remaining messages are considered as non-toxic. \newcite{Karan2019} report an annotation Precision of 51\% for their semi-automated method on a test set of 100 manually annotated messages. Thus, we can suppose that human annotations would surely be more accurate than the semi-automatically generated annotations of \textit{PreTox}.

\subsection{Discussion}
\label{subsec:Discussion}
In this section, we reviewed the corpora related to the detection of abusive messages in Wikipedia talk pages. However, they all have some weaknesses. WCC contains high quality annotations for $3$ types of abusive content, but does not provide any conversational structure. On the contrary, WikiConv provides full conversations but without annotations. PreTox seeks to extend the latter by semi-automatically annotating the messages, but this process is not accurate enough. In the next section, we address these issues by combining WCC and WikiConv to combine their advantages, and thus compensate for their individual drawbacks.

Moreover, we also reviewed the works leveraging these corpora, and highlighted a major issue, as shown in Table~\ref{tab:RecapPerformances}: the lack of a standard protocol for evaluating the performances of abuse detection tools. This flaw concerns both the evaluation metric used and the way the data is divided into train/development/test subsets. In addition, none of the listed works give an open source version of their code. Therefore, it is extremely complicated to have a comparative overview of all proposed approaches, which certainly constitutes a major obstacle to progress in this research area.

\section{Proposed Corpus}
\label{sec:ProposedCorpus}
The context of messages ({\it i.e.} the messages surrounding a targeted message in a conversation) is ignored by many existing abusive content detection methods~\cite{Pavlopoulos2017,Razavi2010,Djuric2015}, while it seems to have a positive influence on classification performances~\cite{Papegnies2019,Yin2009}. An annotated conversation corpus could allow the use of such information at a large scale, in order to develop context-based methods that takes advantage of conversational structure and dynamics to detect abusive content. 

In this work we propose \textit{Wikipedia Abusive Conversations} (WAC), a corpus of messages from Wikipedia integrating conversational information and high quality human annotations. WAC is a combination of the first two corpora described in Section~\ref{sec:RelatedWork}~and takes advantage of their complementarity. It is based on the messages and conversations structure from WikiConv~\cite{Hua2018} and the human annotations for 3 different types of abusive content from the WCC~\cite{Wulczyn2017}. The textual elements constituting conversations in WAC are called ''messages'' like WikiConv as they correspond to WikiConv messages matched with a WCC annotation.
This reconstruction task is not trivial because of the way that comments and messages are identified in the two corpora. As explained before and shown on Figure~\ref{fig:RevidExample}, there is no guarantee of \texttt{rev\_id} uniqueness for WikiConv messages, making it difficult to match them with WCC comments. WAC provides a large collection of conversations including at least one human-annotated message per conversation. It is divided into 3 datasets annotated for \textit{Personal attack}, \textit{Aggression} and \textit{Toxicity}. In total, it contains approximately 193k conversations consisting of 4.9 million messages, among which 383k are annotated. It is publicly available online\footnote{DOI: \href{https://doi.org/10.6084/m9.figshare.11299118}{\texttt{10.6084/m9.figshare.11299118}}}.

\subsection{Reconstruction Pipeline}
\label{subsec:Pipeline} 
\begin{figure*}[!ht]
    \begin{center}
        \includegraphics[width=\textwidth]{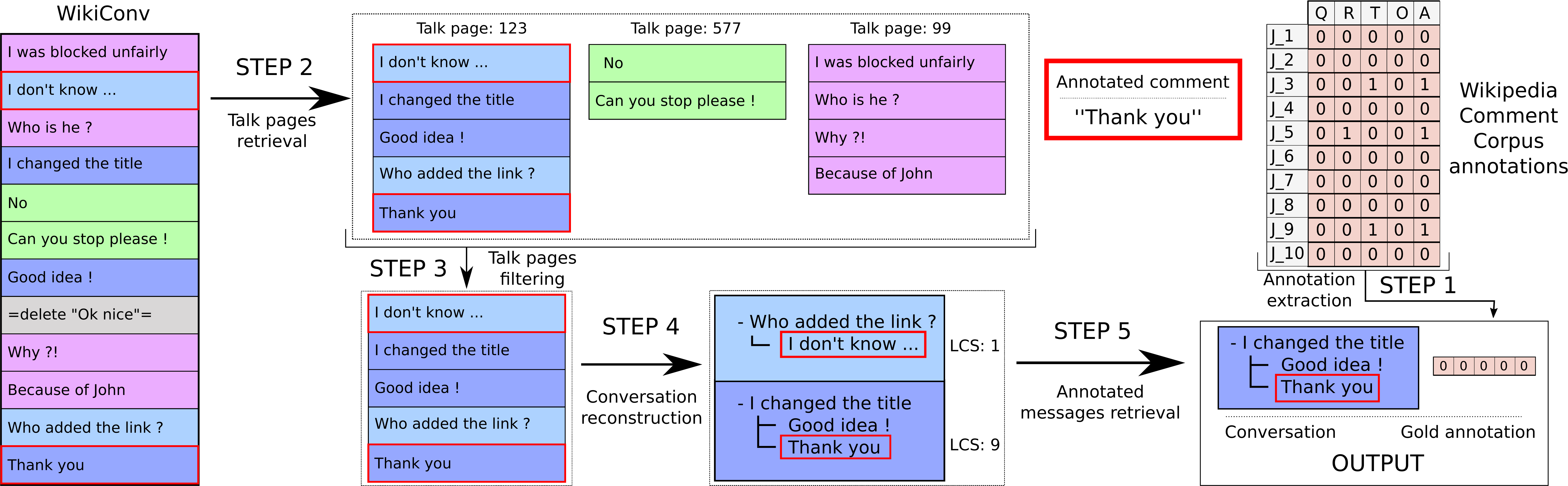} 
        \caption{Representation of our pipeline applied to reconstruct the conversation of an annotated comment. Only the textual content of the actions is displayed. The right side shows the annotated comment and its 10 associated human judgments (J\_1, ..., J\_10). The letters in the WCC table stands for \textit{quoting\_attack} (Q), \textit{recipient\_attack} (R), \textit{third\_party\_attack} (T), \textit{other\_attack} (O), \textit{attack} (A). Messages with a red frame have the same \textit{rev\_id} as the annotated comment. Message colors match both the pages and the conversation containing them. Figure available at \href{https://doi.org/10.6084/m9.figshare.11302385}{10.6084/m9.figshare.11302385} under CC-BY license.}
        \label{fig:ReconstructionPipeline}
    \end{center}
\end{figure*}

We now describe the reconstruction process we developed in order to gather information from existing corpora~\cite{Wulczyn2017,Hua2018} in a new one and extract useful information.
The pipeline is detailed only for the \textit{Personal attack} dataset, but is the same for the other two datasets. Its source code is open source, and publicly available online\footnote{\href{https://github.com/CompNet/WikiSynch}{https://github.com/CompNet/WikiSynch}}.
The reconstruction process is divided into 5 main steps. It begins with the extraction of the annotation from WCC. The second step is to retrieve messages from WikiConv. The third step consists in filtering these messages in order to keep only the relevant talk pages. The fourth step is the conversation reconstruction. The last step, the most important and difficult one, consists in uniquely identifying all the annotated messages in the conversation. Figure~\ref{fig:ReconstructionPipeline} shows the whole pipeline, discussed through this section.

\subsubsection{Annotation Extraction}
The first step is to extract annotations from the WCC. This corpus provides 10 judgments per annotated comment. Each judgement provides multiple annotations depending on the dataset. The \textit{Personal attack} dataset has $5$ binary annotations: \textit{quoting\_attack, recipient\_attack, third\_party\_attack, other\_attack} and the more general \textit{attack}. The \textit{Aggression} and \textit{Toxicity} datasets also provides such a general binary score (\textit{aggression} and \textit{toxicity}, respectively). Additionally, they provide  an \textit{aggression\_score} and a \textit{toxicity\_score}) ranging from $-2$ (very abusive) to $2$ (very healthy), $0$ being neutral.
We aggregate these 10 judgments to determine the gold annotation of all the annotated messages. For the binary annotations, we compute the majority annotation among crowdworkers to determine the gold standard. For the scores, we compute the average value among all crowdworkers.
In WAC, we call \textit{annotated messages} the annotated comments extracted from WCC. 
The right side of Figure~\ref{fig:ReconstructionPipeline} shows an example of this step applied to a comment annotated for \textit{Personal attack}. Based on the 10 human judgments from WCC, the gold annotations for each of the $5$ types of attack annotated is determined.

\subsubsection{Talk Pages Retrieval}
The second step is to retrieve the data from WikiConv\footnote{DOI: \href{https://doi.org/10.6084/m9.figshare.7376003}{\texttt{10.6084/m9.figshare.7376003}}}. Because WCC contains data from the English Wikipedia, we only consider the English part of WikiConv, which is composed of approximately $91$M distinct conversations. We group all messages by their respective \texttt{page\_id}. Note that at this stage, messages are not ordered, and not structured as conversations, as a single talk page can contain multiple conversations. This is for instance the case for the blue talk page in Figure~\ref{fig:ReconstructionPipeline}.
At this step, we also filter the messages based on their type: creations, additions, and modifications are retained while deletions and restorations are filtered out.
Indeed, deletions often concern abusive messages which are already outnumbered by non-abusive messages in WCC, so considering that some messages are deleted would unbalance the corpus even more. By doing this filtering, we also retain a maximum of annotated messages in our corpus.

In the example displayed in Figure~\ref{fig:ReconstructionPipeline}, the colors of the WikiConv messages match the talk page on which they appear. We can distinguish $4$ pages: purple, blue, green and grey. However, the grey page contains only $1$ message which corresponds to a deletion. Thus, this message is removed and after Step 2, only $3$ talk pages remain.

\subsubsection{Talk Pages Filtering}
Among the WikiConv talk pages retrieved at the previous step, only a fraction contains a message that is annotated in WCC. This filtering step aims at keeping only the pages containing at least one such message, in order to retain only the relevant talk pages. However, it is important to understand that the annotated message can be in any of the conversations taking place on the concerned talk page. In order to perform such a filtering, we rely on the \texttt{rev\_id}, the id of the revision from which the message was extracted. The retained pages are all the pages whose at least one message has the same \texttt{rev\_id} as an annotated comment. As previously mentioned, this attribute is available in both WCC and WikiConv but unlike WCC, WikiConv is likely to associate the same \texttt{rev\_id} to several distinct messages (or rather, these messages correspond to a single comment in the WCC). Therefore, there are more messages in WikiConv having the \texttt{rev\_id} of an annotated WCC comment than the actual number of annotated comments. This issue is addressed later at Step~$5$. Messages with a red frame in Figure~\ref{fig:ReconstructionPipeline} are messages having the same \texttt{rev\_id} as the annotated comment. After Step~3, only the blue page is retained, as it is the only one containing messages with the wanted \texttt{rev\_id}. Messages in this page are still unordered.

\subsubsection{Conversation Reconstruction}
\begin{figure}[!ht]
    \begin{center}
        \includegraphics[width=\columnwidth]{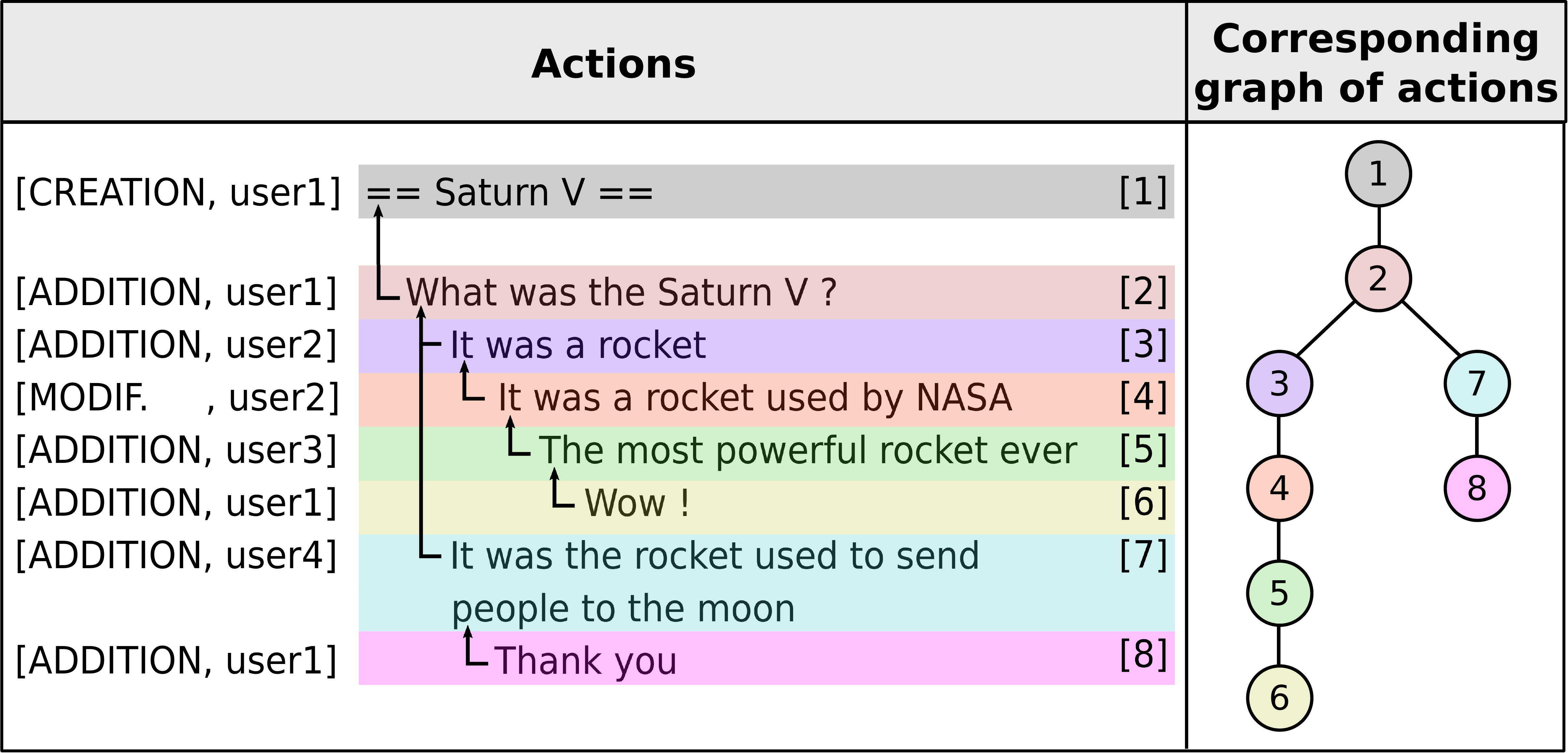} 
        \caption{An example of a conversation and its corresponding graph of actions. Arrows illustrate which action is replying to. Figure available at \href{https://doi.org/10.6084/m9.figshare.11302385}{10.6084/m9.figshare.11302385} under CC-BY license.}
        \label{fig:GraphActions}
    \end{center}
\end{figure}

The fourth step consists in reconstructing the conversation. For each page remaining after the previous step, we reconstruct the distinct conversations taking place on this specific page, using the attributes available in WikiConv.
The reconstruction process starts by retrieving all the messages corresponding to the creation of a new conversation. On Figure~\ref{fig:ReconstructionPipeline}, there are two such messages out of the five messages of this page. To retrieve all the creation of conversations, the \texttt{type} attribute is not enough because some messages being the starting point of a new conversation are categorized as addition and not as creation. 
Then, based on the \texttt{replyTo\_id} allowing to find the message to which this message answers, it is possible to link each message of the conversation and so, to reconstruct its structure. As a result, the structure of each conversation on the page is modeled as a graph of actions, a message being the textual content of an action. Figure~\ref{fig:GraphActions} is an example of a conversation reconstructed during this step. The left part shows the textual content of all the actions in this conversation. The very first action is the creation of the conversation. Then, all the source-reply relationships are modeled using tabulations. An action is a reply to the nearest previous action with one less tabulation. For instance, Actions~7 and~3 are replies to Action~2 which is itself a reply to Action~1. The right part of Figure~\ref{fig:GraphActions} shows the corresponding graph of actions.

During this reconstruction step, modification messages are considered as replies to the original message they are editing in order to keep track of all the content added to the talk page and not only the final form of each message. Moreover, a lot of messages categorized as modifications can actually be considered as additions. Indeed, a typical behavior of Wikipedia users is to reply to a message by adding text straight into the message they want to reply to instead of creating a new one. Thus, some conversations take place in a single message which is modified successively by multiple users. However, even if the conversation takes place in a single message visually, technically, an action is created and saved for each successive edition of the message. So, the graph of action that we produce is exactly the same as if all the messages were posted in successive and distinct messages. This behavior justifies the need to consider modifications as full messages, and not as a simple state of a message at a given time. 

In the example of Figure~\ref{fig:ReconstructionPipeline}, the page retained at the previous step contains $2$ conversations modeled by two distinct shades of blue. These conversations are reconstructed at Step~4 and actions are ordered as they appear on the original talk page.

\subsubsection{Annotated Messages Retrieval}    
We now have reconstructed all the conversations appearing in pages known to contain at least one annotated message. However, some of these conversations may not contain any such message, as several conversations generally coexist on the same talk page. The last step is therefore to filter them out. As stated in Step~$3$, a given \texttt{rev\_id} can be associated to several WikiConv message, whereas it points out at a unique WCC comment. This is a major issue for us because the \texttt{rev\_id} is the only attribute available to match WCC comments to WikiConv messages. In order to figure out which of the messages with equivalent \texttt{rev\_id} is actually the comment annotated in WCC, we compute the \textit{Longest Common Sequence} (LCS) between the original annotated comment and each message in our corpus having the same \texttt{rev\_id}. We consider that the message with the LCS corresponds to the annotated comment. Approximately $36$\% of our annotated messages are concerned by this issue, most of them having only $2$ or $3$ messages with similar \texttt{rev\_id}.

Once every annotated message has been uniquely identified, we filter all the conversations reconstructed at Step~$4$ to only keep those containing at least one annotated message. In the example of Figure~\ref{fig:ReconstructionPipeline}, both conversations reconstructed at Step~4 contain a message with the \texttt{rev\_id} of the annotated comment, the messages with a red frame. The LCS is computed between both conversation messages and the annotated comment. As a result, the message ``Thank you'' is identified as the actual annotated message and its conversation is retained while the other is discarded. In the end, we get the conversation containing the annotated message and its associated gold annotation computed from WCC.

The described pipeline is applied for all annotations types (\textit{i.e., Personal attack, Aggression and Toxicity}) to create the 3 distinct datasets constituting our WAC corpus. It is composed of conversations containing at least one annotated message. Three files containing the annotations are released along with the corpus, each file corresponding to a dataset.

\subsection{Description}
A number of annotated comments from the original WCC datasets are discarded during the reconstruction pipeline described in Section~\ref{subsec:Pipeline}. This is mostly due to missing data in WikiConv or WCC. Some lost comments also are comments associated to a deletion or restoration in WikiConv which are discarded in the reconstruction pipeline. However, $97.97$\% of the original annotated WCC comments are retained in WAC. In total, the corpus contains more than $2.2$ million unique messages split into $168{,}827$ unique conversations. The number of annotated messages and the division of annotations in all three datasets is summarized in Table~\ref{tab:SizeRepartition}. As mentioned in Section~\ref{subsec:WCC}, one annotated message can be annotated for different types of abuse. Hence the $382{,}665$ total annotations from the last line of Table~\ref{tab:SizeRepartition} are assigned to a total of $193{,}265$ distinct messages.

\begin{table}[!ht]
\centering
  \begin{tabularx}{\columnwidth}{ |X|c|c|c| }
    \hline
    \textbf{Dataset} & \textbf{Annotated} & \textbf{Abuse} &
    \textbf{Non-abuse} \\
    & \textbf{messages} & & \\
    \hline
    Personal & 113,174 & 14,934 & 98,240 \\
    attack & & (13.20\%) & (86.80\%) \\
    \hline
    Aggression & 113,174 & 16,331 & 96,843 \\
    & & (14.43\%) & (85.57\%) \\
    \hline
    Toxicity & 156,317 & 19,700 & 136,617 \\
    & & (12.60\%) & (87.40\%) \\
    \hline\hline
    Total & 382,665 & 50,965 & 331,700 \\
    & & (13.31\%) & (86.69\%) \\
    \hline
  \end{tabularx}
  \caption{Number of annotated messages and distribution of annotations in the proposed \textit{Wikipedia Abusive Conversations} (WAC) corpus.}
  \label{tab:SizeRepartition}
\end{table}

\begin{figure}[!ht]
    \begin{center}
        \includegraphics[width=\columnwidth]{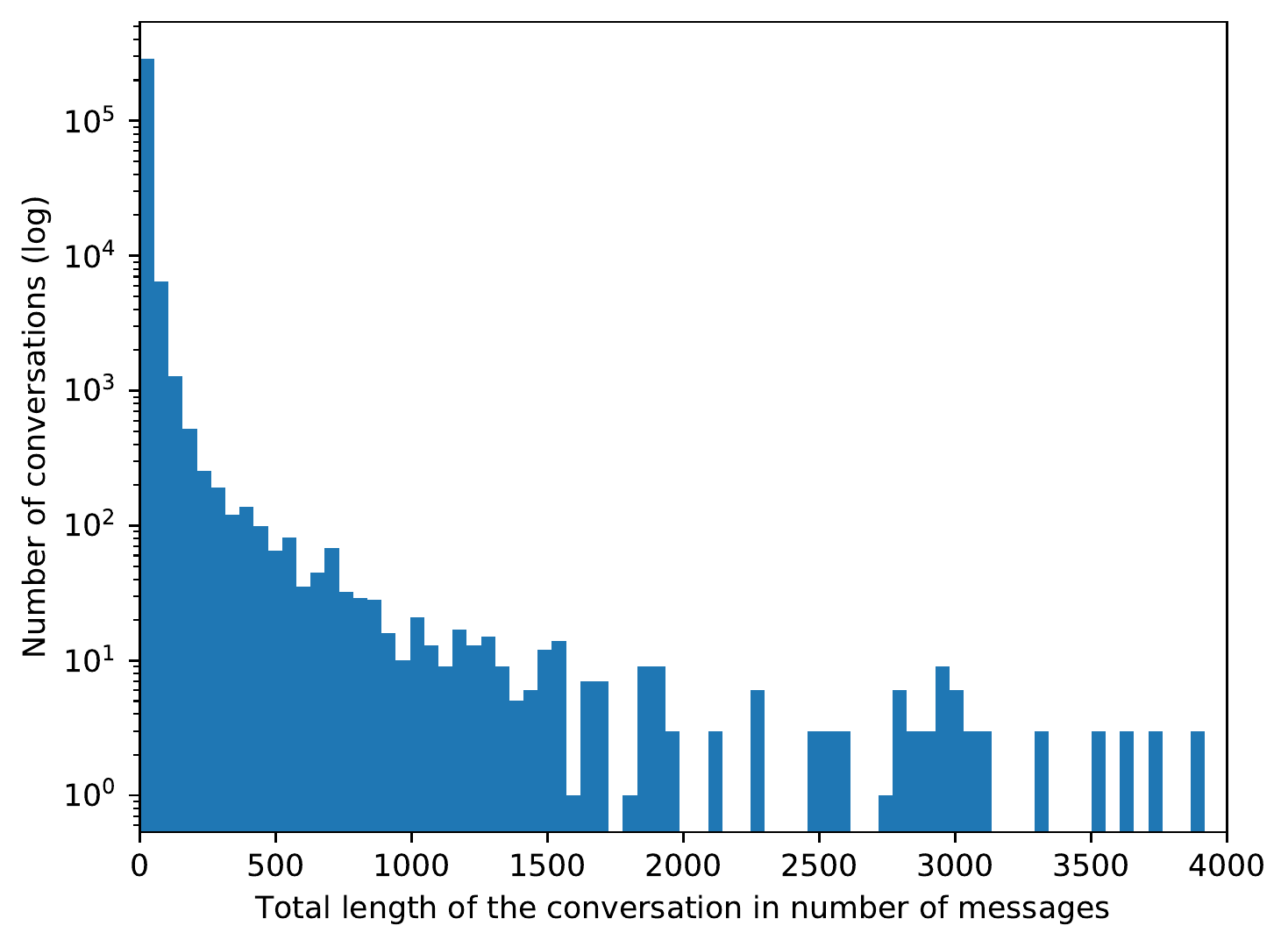} 
        \caption{Distribution of the conversation lengths in \textit{Wikipedia abusive Conversations}, expressed in number of messages. The \textit{y}-axis scale is logarithmic. Figure available at \href{https://doi.org/10.6084/m9.figshare.11302385}{10.6084/m9.figshare.11302385} under CC-BY license.}
        \label{fig:ConvLength}
    \end{center}
\end{figure}

Wikipedia messages are usually longer than other types of online posts such as tweets or chat messages. Messages in our corpus have an average length of more than $1{,}000$ characters. 
As shown in Figure~\ref{fig:GraphActions}, the structure of each conversation can be modeled as a graph. On average, there are $13$ messages in a conversation. The distribution of the conversations length is shown in Figure~\ref{fig:ConvLength}. Note that the \textit{y}-axis scale is logarithmic for readability reasons. We can observe that some conversations contain more than $1{,}000$ messages but for a large majority, conversations are only $1$- to $20$-message long.

\begin{figure}[!ht]
    \begin{center}
        \includegraphics[width=\columnwidth]{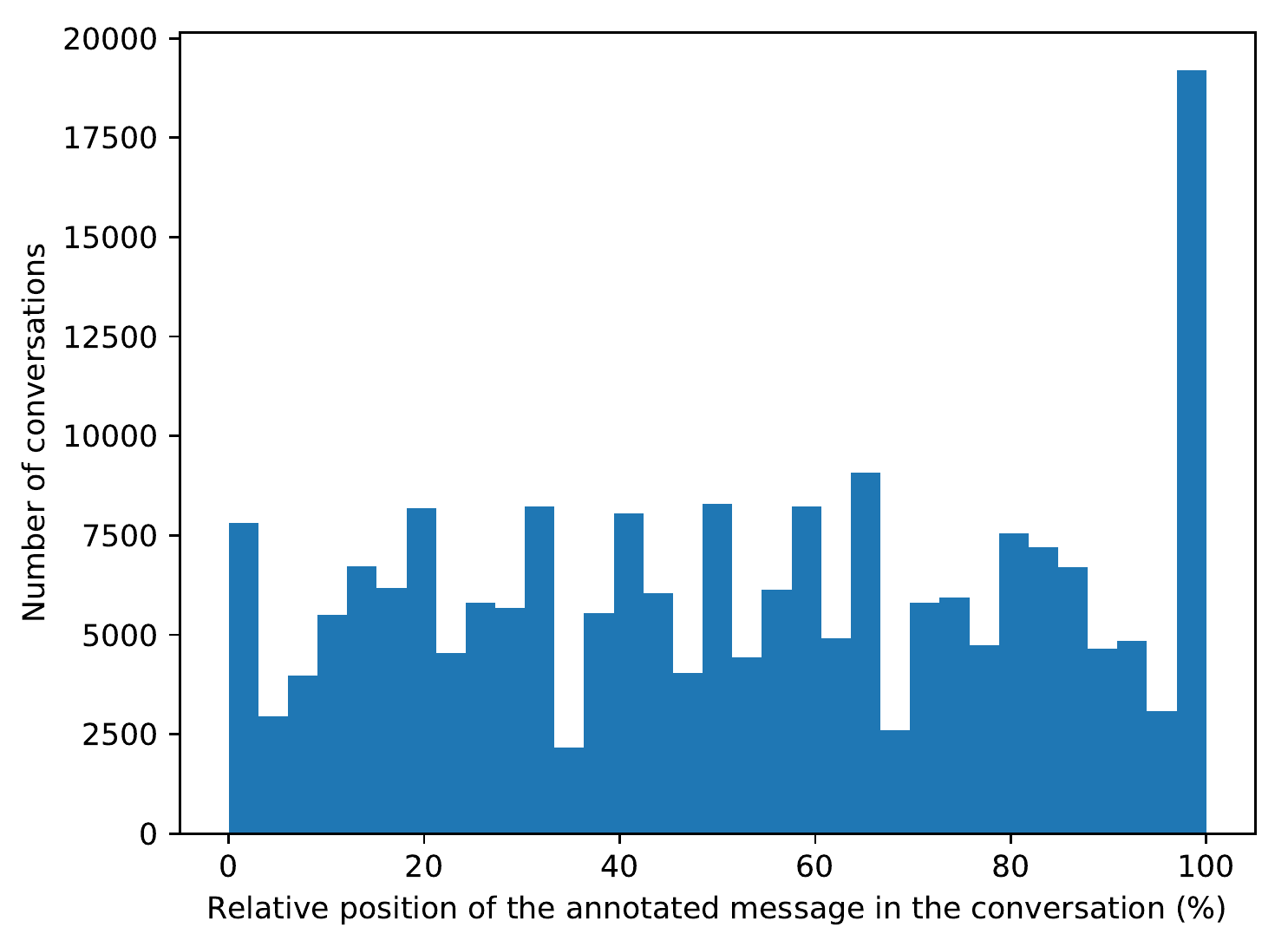} 
        \caption{Position of the annotated messages in the conversation, expressed in percentage of the messages appearing \textit{before} the annotated message. Only conversations with $5$ and more messages are considered. Figure available at \href{https://doi.org/10.6084/m9.figshare.11302385}{10.6084/m9.figshare.11302385} under CC-BY license.}
        \label{fig:AnnotatedPos}
    \end{center}
\end{figure}

Figure~\ref{fig:AnnotatedPos} shows the distribution of the relative position of annotated messages in conversations. This position is expressed as the percentage of messages posted \textit{before} the annotated message in the conversation. Only conversations with at least $5$ messages are considered in this figure. This distribution shows that annotated messages are well distributed over all positions in the conversations, except at the very end of the conversation where a lot more annotated messages appear. This observation holds whether the annotated message is abusive or not. For abusive messages, this position can potentially be explained by the fact that abusive comments are quickly deleted from Wikipedia~\cite{Hua2018}, before creating many reactions.

Conversations can also be divided into multiple sub-conversations, a sub-conversation being a sequence of messages in which each message is an answer to the previous. For instance, the conversation represented in Figure~\ref{fig:GraphActions} is composed of $8$ messages and $2$ sub-conversations. On average, there are $3$ sub-conversations per conversation in the corpus. However, they only contain approximately 4 messages in average, which is quite limited for a conversation. 
This value highlights the fact that Wikipedia talk pages are not used in the same way as forums or social media. Indeed, many messages are informative messages explaining what changes have been made to the article or suggestions on how to edit the article associated to the talk page. Most of the time, these messages do not imply answers, which can explain the relatively small number of messages per sub-conversation that we observe.

\section{Proposed Benchmarking Platform}
\label{sec:ProposedBenchmark}
In Section~\ref{subsec:Discussion}, we highlighted some issues with the evaluation and comparison of current abuse detection methods. To overcome this problem, we propose a common benchmark platform, described in Section~\ref{s:plateform}. Then, we assess, in Section~\ref{s:usage}, existing detection methods to illustrate the interest of our corpus and platform.

\begin{table*}[!htb]
\centering
  \begin{tabular} { |p{3cm}|r|r|r|r|r|r|r|r|r| }
    \hline
     \textbf{Ground Truth} & \multicolumn{3}{c|}{\textbf{Perspective - Toxicity}} & \multicolumn{3}{c|}{\textbf{Perspective - Severe toxicity}} & \multicolumn{3}{c|}{\textbf{Hybrid method}} \\
    \cline{2-10}
    \textbf{Dataset} & Prec. & Recall &
    $F$-measure & Prec. & Recall &
    $F$-measure & Prec. & Recall &
    $F$-measure \\
    \hline
    Personal attack & 84.96 & 86.96 & 85.96 & 54.05 & 93.06 & 68.38 & 81.24 & 70.47 & 75.47 \\
    \hline
    Aggression & 82.89 & 87.49 & 85.13 & 53.71 & 92.52 & 67.96 & 81.75 & 70.23 & 75.55 \\
    \hline
    Toxicity & 84.68 & 90.82 & 87.64 & 53.49 & 94.29 & 68.26 & 74.17 & 74.77 & 74.47 \\
    \hline
  \end{tabular}
  \caption{Macro Precision, Recall and $F$-measure obtained by the $3$ tested methods. }
  \label{tab:PerfsResults}
\end{table*}

\subsection{Platform Description}
\label{s:plateform}

An important issue is that the reported performance in different works are often almost impossible to compare since systems may be evaluated on different datasets, and many different metrics are used to perform these evaluations. Even if the used corpus is identical, which is the case with works on WCC for example, the way it is split into train/development/test differs, and is often not precisely described by the authors. 

All these issues hinder replicability. In this section, we present a benchmarking platform that we developed in order to address them. It is an open-source tool available online\footnote{\href{https://github.com/CompNet/Alert}{https://github.com/CompNet/Alert}}, and aimed at grouping classification methods in the area of automatic abusive content detection to ease and stimulate the replicability of the reported performances. Moreover, we take advantage of our new corpus to address the difficulties in the comparison of the results. All the methods of the platform are assessed using the corpus we developed (WAC). We propose a split into train (60\%), development (20\%) and test (20\%) sets for each of the 3 datasets of WAC. This split was randomly generated, but is publicly available online. Using this split for all the methods ensure that all the results are obtained with the same data and so, are truly comparable. Additionally, we leave open the possibility to implement and add further metrics to the methods if needed, the tool being designed to ease the addition of new metrics. Different variants of the $F$-Measure as well as the Area Under the ROC Curve are currently implemented, since they are the metrics mainly used by the methods listed in Table~\ref{tab:RecapPerformances}.

As mentioned before, the source code of all the works presented in Section~\ref{sec:RelatedWork} is not publicly available, so we could not include them in our platform. Instead, we focused on our previously published abuse detection method~\cite{Cecillon2019}. It is a hybrid approach combining two distinct methods previously proposed by our team~\cite{Papegnies2019}. The first one is content-based and relies on a set of features describing exclusively the textual content of the messages to perform the classification. Though this approach does not require a corpus of conversations to be tested, it is still interesting to assess it on the WAC corpus because of its size. Indeed, the reported performances for this method were obtained on a small dataset of less than $3{,}000$ comments. 
The second approach is graph-based, and requires full conversations in order to be applied. It consists in extracting conversational networks modeling the interactions between users, before computing topological measures to describe these graphs, and using them as features during the classification process. Hence, this method completely ignores the content of the messages and only relies on the structure and dynamics of the conversation. It is typically the kind of methods for which the WAC was created. Our hybrid tool combines both text- and graph-based features.

Our benchmarking platform is currently available online, but still needs some work to implement more existing approaches. Indeed, while it currently contains only two methods, the objective is to include more methods using \textit{Wikipedia conversations} to make it a comparative platform of many approaches in the area of automatic abuse detection.

\subsection{Usage Example}
\label{s:usage}

We now illustrate the interest of our corpus and platform to assess the performance of the Google Perspective API as well as our own hybrid method. The way the data is split between train, development, and test sets is described in a file provided with the data. We assess the performance in terms of Precision, Recall and $F$-measure, separately for the $3$ datasets constituting WAC ({\it Personal attack}, {\it Aggression} and {\it Toxicity}). Our results are presented in Table~\ref{tab:PerfsResults}. 


Each message in WikiConv is provided with two scores computed through the Google Perspective API: a \texttt{toxicity} and a \texttt{severe\_toxicity}. To evaluate their quality, we first convert them into binary classes (\textit{abusive} vs. \textit{non-abusive}), by using the equal error thresholds calculated by \newcite{Hua2018} following the methodology of \newcite{Wulczyn2017}. A message is considered toxic if its \texttt{toxicity} score is above $0.64$ and severely toxic if its \texttt{severe\_toxicity} score is above $0.92$. Unsurprisingly, the best $F$-measure for the \texttt{toxicity} score is obtained with the \textit{Toxicity} dataset. However, the performances obtained for the other $2$ datasets are not much lower. Based on this observation, we can hypothesize that the method used to generate the \texttt{toxicity} and \texttt{severe\_toxicity} scores may not really distinguish between \textit{Personal attack}, \textit{Aggression} and \textit{Toxicity}, and relies on a more general definition of abuse. The \texttt{severe\_toxicity} score yields a higher Recall than the \texttt{toxicity} one for all $3$ abuse types, but the Precision is only around 54\%. This poor precision is due to a lot of toxic messages being mistaken for severely toxic messages. This confirms our assumption from Section~\ref{sec:ProposedCorpus}, \textit{i.e.} \textit{PreTox} annotations (largely based on the Google Perspective API) are less accurate than human annotations.

For the hybrid method implemented in our platform, performances are similar for the $3$ datasets, with a $F$-measure around 75\%. This puts our method between both variants of the Google Perspective API. There is a clear drop in performance compared to the results obtained in our previous work, on a different corpus~\cite{Cecillon2019}. This can be explained by several factors. First, the text-based part of our method relies on very standard features and could be improved by using more sophisticated ones. Second, the graph-based part was designed to operate on chat messages, and therefore to handle very large and linear conversations. In WAC, conversations have a limited size and are not linear, which decreases a lot the efficiency of this method. Therefore, there is room for improvement, and we plan to adjust both parts to better handle the characteristics of Wikipedia talk pages. In any way, our goal in this paper was only to illustrate the usefulness of our platform and corpus, and we leave the improvement of our classifier to future work.

\section{Conclusion and Future Work}
\label{sec:Conclusion}
In this paper, we introduced a large corpus of 383k annotated user messages along with the conversations they appear in. We presented the pipeline that we developed to link two existing corpora of Wikipedia comments and extract high quality labels and thread-level information. 
So far, the development of context-based methods in the area of abusive comment detection was limited by the lack of large annotated corpora of conversations. This new publicly available corpus opens perspectives for new work and for extending existing work. For example, content-based methods could  incorporate information about the conversation and its structure. Furthermore, the large number of messages in the corpus allows us to use it with any machine learning approach.
In a second part, we presented a tool that we developed to assess some detection methods on this new corpus. A future work is to further develop this platform by integrating more methods in it. The objective is to make it a comparison platform for classification methods using the conversational corpus we proposed.


\section{Bibliographical References}
\label{main:ref}
\bibliographystyle{lrec}
\bibliography{biblio}


\end{document}